\pdfoutput=1

\documentclass[11pt]{article}

\usepackage[preprint]{acl}

\usepackage{times}
\usepackage{latexsym}
\usepackage{paralist}

\usepackage[T1]{fontenc}

\usepackage[utf8]{inputenc}

\usepackage{microtype}

\usepackage{inconsolata}

\usepackage{graphicx}

\newcommand{\para}[1]{\vspace{3pt} \noindent {\bf #1}}

\title{On Defining Erasure Harms for NLP}

\author{\textbf{Yu Lu Liu}$^{1}$\thanks{This work was partially supported by and started as part of a Microsoft Research - Mila collaboration and grant.} ~~~~\textbf{Arnav Goel}$^{2}$~~~~\textbf{Jackie Chi Kit Cheung}$^{3,4,5,*}$
\\
\textbf{Alexandra Olteanu}$^{3, *}$~~~~\textbf{Ziang Xiao}$^{1}$~~~~\textbf{Su Lin Blodgett}$^{3, *}$ \\
    $^1$Johns Hopkins University \quad
    $^2$Carnegie Mellon University \\
    $^3$Mila -- Qu\'{e}bec Artificial Intelligence Institute \quad
    $^4$McGill University \\
    $^5$Canada CIFAR AI Chair, Mila \\
    \texttt{yliu624@jh.edu} \quad
    \texttt{ziang.xiao@jhu.edu} 
}

\usepackage{comment}
\usepackage{todonotes}
\usepackage{rotating}
\usepackage{booktabs}
\usepackage[table]{xcolor}

\begin{document}
\maketitle
\begin{abstract}
The deployment of NLP systems has raised concerns about harms they might produce, including representational harms. Recent literature has begun to conceptualize and measure one such harm, the harm of \emph{erasure}. Nevertheless, the field lacks a clear and cohesive conceptual foundation for identifying and measuring erasure. Existing conceptualizations of erasure are often {\em broad}---making it difficult to identify what is needed to establish and measure erasure---or else {\em specific} to particular settings---facilitating measurement for those settings but potentially challenging to adapt to other settings. To address this gap, 
we develop and propose a {\em structured definition} of erasure that clarifies what components are necessary for establishing whether erasure has occurred, which practitioners need to explicitly articulate and operationalize in order to measure erasure.\looseness=-1 
\end{abstract}

\section{Introduction}
\vspace{-4pt}
The deployment of natural language processing (NLP) systems has led to concerns about the harms these systems might produce. A significant body of existing literature focuses on conceptualizing, measuring, and mitigating representational harms such as stereotyping \cite[e.g.,][]{caliskan2017semantics,cao-etal-2022-theory,kotek2023gender} or hateful or toxic speech \cite[e.g.,][]{rottger-etal-2021-hatecheck,gehman-etal-2020-realtoxicityprompts} produced or reproduced by NLP systems. 
Although \textit{erasure} has been recognized as a type of representational harm \cite{SuLin_Thesis}, it is only recently that the NLP research community has begun showing interest in conceptualizing and measuring it \cite[e.g.][]{dev-etal-2021-harms,katzman2023taxonomizing,schwobel-etal-2023-geographical, qadri2025risksculturalerasurelarge}. \looseness=-1

\begin{figure}[t!]
    \centering
    \includegraphics[scale = 0.75]{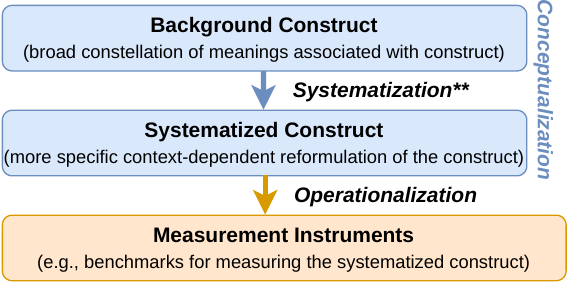}
    \caption{Illustration of conceptualization and operationalization adapted from \citet{wallach2025evaluating}. Our work focuses on systematization, marked by **.\looseness=-1} 
    \vspace{-10pt}
    \label{fig:conceptualization_diagram}
\end{figure}

At the same time, {\em constructs like ``erasure'' are unobservable}, meaning that their presence can only be inferred from what is observable (e.g., in LLM outputs). Because of this, it can be hard to achieve conceptual clarity, which is critical to the creation and use of {\em valid} measurement instruments (e.g., metrics, benchmark datasets). 
In addition, developing measurement instruments that meaningfully capture constructs of interest also requires that those constructs are sufficiently specific---e.g., grounded in specific contexts of system development, deployment, or evaluation.
Practitioners working in different contexts (e.g., different cultural contexts) are also likely to have differing views on how a construct like ``erasure'' should be defined. \looseness=-1 
This means that existing measurement instruments may be difficult to use as-is across contexts, and their results difficult to interpret, particularly if debates about conceptual differences and the development of instruments are not supported by a cohesive conceptual foundation.

\begin{figure*}[t!]
    \centering
    \includegraphics[scale = 0.73]{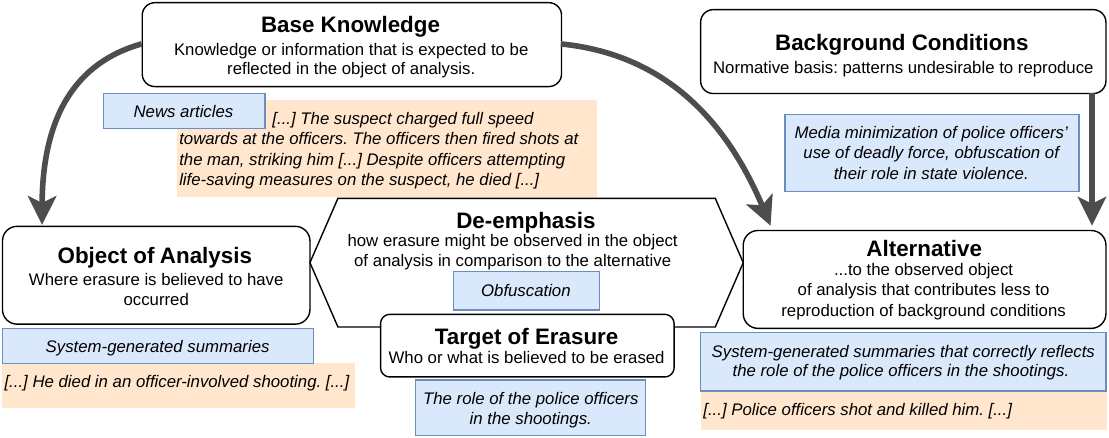}
    \vspace{-10pt}
    \caption{Overview of the six determinant components in our definition of erasure. To systematize erasure to a particular context, practitioners specify these components (in blue) by e.g., considering a specific system output (in orange). This example examining a news summarization system is discussed in more detail in Section~\ref{sec: apply_def}.}
    \vspace{-10pt}
    \label{fig:erasure_diagram}
\end{figure*}

In both NLP and machine learning (ML), such concerns about existing measurement instruments have resulted in their validity being increasingly questioned~\cite[e.g.,][]{blodgett-etal-2021-stereotyping,jacobs2021measurement,wallach2025evaluating}.
These critiques often draw on work in the social sciences, which have a long history of seeking to measure unobservable and often contested constructs~\cite{adcock_collier}.\looseness=-1

In particular, \emph{measurement theory}, which studies how to develop appropriate measures of desired constructs \cite{measurement_theory_2018,jacobs2021measurement}, explicitly distinguishes \emph{conceptualizations}---what we want to measure---from \emph{operationalizations}---how we measure it, and establishes clear, well-grounded conceptualizations as a prerequisite for measurement.
To obtain such conceptualizations, \citet{adcock_collier} also distinguish between a \emph{background construct}---``the constellation of potentially diverse meanings associated with a given [construct]''---and a \emph{systematized construct}, a narrower, more context-dependent version of a background construct. \citet{wallach2025evaluating} describe the process of \emph{systematization} as the movement from a background construct to a systematized one, which in turn enables operationalization. 
Practitioners face many complex choices in this process, but it ultimately allows them to fruitfully proceed with developing measurement instruments, and later on applying and interrogating them.\looseness=-1
\footnote{For clarity and consistency, throughout this paper we use the term ``construct,'' whereas \citet{adcock_collier} instead use the term ``concept.''} 

In the existing NLP literature on establishing and measuring erasure harms, practitioners have systematized {\em erasure harms} to various degrees. Much of this work proposes or relies on definitions of the construct that are either broad---e.g., \emph{``lack of adequate representation of members of a particular social group''} \cite{dev-etal-2022-measures}---lacking a clear way for practitioners to identify what is needed to establish and measure erasure---or else already systematized with respect to specific settings---e.g., image tagging \cite{katzman2023taxonomizing}---facilitating measurement for those settings but potentially challenging to adapt to others, including because it might be unclear how different systematized versions of the construct relate to each other. Thus far, {\em little work has focused on the process of systematization}.\looseness=-1 

To address this gap and facilitate the process of systematization~(Figure~\ref{fig:conceptualization_diagram}),  \emph{we propose a structured definition of erasure that makes explicit the necessary components that any systematized version of the construct needs to specify}. Our proposed definition, which draws on scholarship in both the social sciences and erasure in NLP, is intended to provide a clear foundation for practitioners seeking to systematize erasure by clarifying what components are {\em required} in order to establish whether erasure has occurred: who or what is being erased, where the erasure is observed, and how we know that erasure has occurred (Figure~\ref{fig:erasure_diagram}). Practitioners can draw on this foundation, and their particular use contexts and normative concerns, to specify these components and thus develop their measurement instruments.
We also showcase how our definition can be used as an analytical framework to map current conceptualizations of erasure, surfacing trends and gaps in how erasure is currently conceptualized and operationalized in NLP.\looseness=-1

\section{Background \& Related Work}
\label{sec: related_work_outsideCS}
\vspace{-4pt}
\subsection{Erasure outside NLP}
\vspace{-4pt}
Scholarship across fields ranging from linguistics and linguistic anthropology to gender, critical race, and Indigenous studies has illustrated {\em how language is implicated in erasure}. 
First, language can be involved in the erasure of a wide range of objects. 
It can be used to erase aspects of people's identity, and can also be involved in the erasure of ideas, topics, and perspectives---e.g., discourses of colorblindness that invisibilize racial injustice~\cite{bonilla2021racism}---and of language varieties---e.g., assimilation policies that sought to suppress Indigenous languages and cultures~\cite{reyhner2010indigenous}.\looseness=-1

Second, language can be involved in erasure via many mechanisms. 
Erasure may happen via different levels of emphasis produced by certain linguistic constructions; e.g., \citet{moreno2022officer} describe ``semantic structures that obfuscate responsibility'' in media writing about police killings, which reduce focus on police violence. 
Speakers can use language to introduce inaccurate information, such as when the print media labelled Olympic diver Tom Daley as gay rather than as bisexual, thereby erasing his orientation\footnote{Limited to the authors' interpretation of Daley's public statements at the time. His actual orientation may differ.}~\cite{bisexual_erasure}.
Speakers may avoid using some language; for example, \citet{hamilton2020whats}'s analysis of corporate statements on racial injustice found that ``companies were reluctant to even use the word `race.'''
Policies and practices may limit discussion or transmission of certain knowledge or topics; for example, 
strategies intended to prevent technologies from producing harmful speech, such as blocklists, can prevent speech about topics like race~\cite{schlesinger2018lets}.\looseness=-1

What \emph{work} does erasure perform? By making claims of what exists or does not exist, or what is (not) possible, erasure often reproduces and legitimizes unjust social arrangements. 
The erasure of individuals or groups can deny their existence, their experiences and histories, or ability to assert their identities and futures---thereby casting them as unworthy of moral or political concern or self-determination \cite{fraser2008abnormal,roche_articulating_2019}, and casting unjust treatment as legitimate. 
The erasure of ideas, topics, and perspectives can make them unavailable for public discourse, ``shap[ing] how [people] understand what's happening and which solutions sound appropriate''~\cite{hamilton2020whats}---e.g., by positioning race (rather than racism) as the source of racial injustice~\cite{hamilton2020whats}.\looseness=-1

Collectively, these insights shape our thinking in a few ways. First, NLP systems necessarily involve decisions about what knowledge to represent and how to represent it (often determined by the assumptions developers and practitioners operate under), and how those decisions are communicated via language.
In developing our definition, we thus sought to encompass the many possible objects that NLP systems might erase and the many mechanisms by which they might do so through language.  
These insights also draw our attention to the \emph{impacts} and \emph{patterns}, historical and ongoing, of erasure---i.e., who is harmed, how beliefs about the world are constructed and reshaped, and what social arrangements are reproduced. Thus, our definition foregrounds social and historical context as critical for establishing that erasure has occurred.\looseness=-1

\subsection{Erasure in NLP} 
\vspace{-4pt}

Some of the NLP literature on biases and harms addresses erasure. 
For example, \citet{dev-etal-2022-measures} propose a framework of harms in NLP to guide practitioners in the development of bias measures. Within this framework, they define ``erasure'' as \textit{``the lack of adequate representation of members of a particular social group [...] whether intentional or not.''} \citet{gallegos-etal-2024-bias} survey bias evaluation and mitigation techniques for LLMs, and define ``erasure'' as \textit{``omission or invisibility of the language and experiences of a social group''} within their taxonomy of social biases. 

There is also emerging work in NLP that systematizes or operationalizes erasure. 
\citet{dev-etal-2021-harms} examine harms of gender exclusivity in language technologies, where erasure is defined as \textit{``the accidental or intentional invalidation or obscuring of nonbinary gender identities \& how stereotypes about nonbinary communities are portrayed and propagated.''} \citet{schwobel-etal-2023-geographical} study ``geographic erasure,'' examining the frequency of geographic regions produced in generated text, while \citet{qadri2025risksculturalerasurelarge} study ``cultural erasure,'' examining how different cultures are represented in location descriptions and travel recommendations. \citet{devinney-etal-2024-dont} study erasure of identity in narratives generated by LLMs, and \citet{venkit2025taleidentitiesethicalaudit}
study erasure in the use of LLMs for generating synthetic personas. 
\citet{Mollema2025} draws from the fields of philosophy of technology, political philosophy, and social epistemology to conceptualize ``generative hermeneutical erasure'': erasure of certain ways of knowing by generative AI. 

\citet{corvi-etal-2025-taxonomizing} employ speech act theory to provide systematized constructs for representational harms, including erasure. Representational harms are conceptualized as \textit{``perlocutionary effects, (i.e., real-world impacts) of particular types of illocutionary acts (i.e., system behaviors).''} They define an \textit{``erasing illocutionary act to be an illocutionary act that invokes within-hierarchy similarity to disempower one or more social groups (or one or more individuals based on their membership [in] those social group(s)).’’} Though they share our motivation of facilitating measurement of erasure, the construct is already systematized to be about social groups; their focus is on providing an already systematized construct.
By contrast, we focus on facilitating the {\em process} of systematization by clarifying which components practitioners need to specify in order to systematize erasure in the context of their own settings and normative concerns.\looseness=-1 

\section{Constructing Our Definition}
\vspace{-2pt}
\label{sec: def_methodology}
To construct our definition, we first identified key components the definition should capture. To do so, we reviewed a purposive sample of literature on erasure (see below), surfacing recurring (and missing) components from how erasure has been understood and conceptualized. We also conducted multiple collaborative discussion sessions during which we considered various NLP tasks, how the components we surfaced would apply for each task, and whether additional components were necessary. 
Finally, we synthesized the identified components to form our proposed definition (\textsection\ref{sec: components_of_erasure}).\looseness=-1

\para{Surfacing Components of Erasure.}
To identify how erasure has been understood and conceptualized, we reviewed a \textit{purposive sample} of existing NLP work on erasure\footnote{We focused on papers using the term ``erasure'' to describe a construct of interest, but excluded papers that did not use the term to refer to a normative concern. For example, we excluded work focusing on machine unlearning where some set of information is to be ``erased’’ from ML model weights.} \cite{dev-etal-2021-harms,dev-etal-2022-measures,schwobel-etal-2023-geographical,gallegos-etal-2024-bias, qadri2025risksculturalerasurelarge}, as well as from other fields such as machine learning and human-computer interaction \cite{misgendering_machine,schlesinger2018lets,katzman2023taxonomizing}, media communication \cite{bisexual_erasure,hamilton2020whats,moreno2022officer}, linguistic anthropology \cite{reyhner2010indigenous, roche_articulating_2019}, sociology \cite{bonilla2021racism}, philosophy \cite{fraser2008abnormal}, and law \cite{dyal2021autocorrecting}.
We examined how these studies have defined or operationalized the construct.
From these existing conceptualizations of erasure, we surfaced recurring components in how erasure is conceptualized; e.g., the importance of social and historical context.
We also took note of components that might help make some of these existing conceptualizations more complete, as well as components we judged to be underlying some of the existing work without having been made explicit in that work.\looseness=-1

\para{Grounding in NLP.}
We held multiple discussion sessions in order to ground the components we surfaced in the context of NLP tasks. To support this process, we gathered examples of NLP system outputs  (Appendix~\ref{appendix: examples}) and deliberated about whether we could establish that erasure occurred in those outputs. 
For instance, we considered the task of automatic text summarization, where summaries are generally expected to only contain information present in the input documents. 
Practitioners may thus find it unreasonable to claim that erasure---e.g., of national identities---occurred in a system's output summaries if people's nationalities were not mentioned in the input documents in the first place. 
These expectations, which may come from task specification or constraints, need to be accounted for in the definition of erasure for NLP.\looseness=-1

\para{Synthesizing and Connecting Components.}
Our discussion sessions subsequently focused on synthesizing the components we surfaced. We grouped those that accomplished similar roles, for example grouping ``social/historical context'' with ``past patterns of harmful NLP system behaviors'' as they all share the role of forming the normative basis upon which we might determine whether erasure occurred. Finally, we connected components based on how they inform or interact with each other. This process yielded six components.\looseness=-1

\section{Proposed Components of Erasure}
\label{sec: components_of_erasure}
We identified six components that practitioners need to specify when systematizing erasure in NLP settings: 
1) the \textbf{target of erasure}: \textit{who} or \textit{what} is believed to be erased; 
2) the \textbf{object of analysis}: \textit{where} erasure is believed to have occurred (e.g., system outputs);
3) the \textbf{background conditions}: social, historical, or past patterns that practitioners judge to be undesirable to reproduce, and which form the normative basis for the analysis of whether erasure has occurred; 
4) the \textbf{base knowledge}: knowledge or information that is expected to be reflected in the object of analysis; 
5)~an \textbf{alternative} 
to the {\em observed} object of analysis that contributes less to the reproduction of background conditions; and
6)~\textbf{de-emphasis}: \textit{how} erasure might be observed in the object of analysis in comparison to the alternative. Our definition connects these components:\looseness=-1
\vspace{-4pt}
\begin{quote}
    \underline{Definition}: Erasure is a \textbf{de-emphasis} of a \textbf{target}---where this de-emphasis is occurring in an \textbf{object of analysis} that is intended to reflect some \textbf{base knowledge}---with respect to an \textbf{alternative} that reproduces less a set of \textbf{background conditions} than the object of analysis does.\looseness=-1
\end{quote}
\vspace{-6pt}

\subsection{What Is Believed to Be Erased and Where}
\label{what-where}
\vspace{-5pt}
\para{Target of erasure.} 
Often, the first consideration when trying to establish whether erasure has occurred is {\em who} (or what aspects related to them) or {\em what is believed to be erased}. 
As highlighted earlier (\S\ref{sec: related_work_outsideCS}), the target of erasure can take many forms, such as attributes related to people's or groups' identities (e.g., gender identity~\cite{misgendering_machine}), as well as language varieties (e.g., minoritized language varieties~\cite{haque2015indigenous,roche_articulating_2019}), topics, perspectives, and histories (e.g., histories or collective experiences of colonialism~\cite{davis2017resisting} or racial injustice~\cite{hamilton2020whats}). We note that there is often no clear delineation between attributes related to people's or groups' identities and objects such as language varieties, topics, or histories, which are often linked to or constitutive of identity~\cite[e.g.,][]{lucy2024one}. 

\para{Object of analysis.}
Another key consideration is where erasure is believed to have occurred or could be observed---i.e., {\em what is analyzed or observed to determine whether erasure has occurred}. 
While we recognize that erasure harms can also be a result of the process of developing and deploying of NLP systems, or a result of which systems are or are not built, in this paper, we focus on settings where practitioners are tasked with determining whether a system gives rise to erasure harms from observing a given NLP system's outputs or behavior (e.g., classification labels, generated text).\looseness=-1 

\subsection{Expectations about the Outcome}
\label{how}
The term ``erasure'' commonly entails removal or absence,\footnote{The dictionary definitions of ``erase'' are ``to remove from existence or memory'' \cite{Merriam-Webster}; ``to scrape or rub out (anything written, engraved, etc.); to efface, expunge, obliterate''~\cite{Oxford-English-Dictionary}.} which then requires a counterfactual---a state where what was removed is present. 
It is thus difficult to determine that erasure has occurred in the object of analysis without having, implicitly or explicitly, an idea of a ``better,'' counterfactual alternative to what is being observed.\looseness=-1  

For example, one may think that in a good meeting summary the contribution of every employee is mentioned,
and by \textit{comparison}, a summary that omits the contribution of some employee might be a case of erasure. 
But how exactly are such alternatives constructed, and what does ``better'' mean? The following components are intended to help answer these questions,  making this comparison and its underlying normative foundation explicit.

\para{Background conditions.} 
The process of determining whether a target was erased---i.e., the comparison between the observed object of analysis and a counterfactual---is inherently \emph{normative}. It requires making the normative judgment that the {\em observed} object of analysis is ``worse'' than some {\em alternative} that is ``better.'' 
To gain clarity on exactly what ``better''/``worse'' could mean across contexts, 
we turn to prior work studying erasure in other fields (\textsection\ref{sec: related_work_outsideCS}), where establishing whether erasure has occurred often requires taking into account social and historical context, as erasure often reproduces unjust social and historical patterns. We refer to these patterns, as well as any other patterns that may be considered undesirable to reproduce, as \textit{background conditions}, which serve as the normative basis on which we analyze whether erasure has occurred: given a set of background conditions that ought not to be reproduced, \textit{an alternative is ``better'' when it does not reproduce these background conditions, or does so to a lesser extent.}\looseness=-1 

We note that people may experience harm even when the chosen target of erasure is not directly or explicitly related to them---e.g., rhetorics that minimize histories of colonial action by omitting mentions of colonial nation states harm not those states whose actions are de-emphasized, but Indigenous communities onto whom such rhetorics ``misattribut[e] agency''~\cite{davis2017resisting}. 
\looseness=-1

\para{Base knowledge.} 
Apart from the normative expectation set up by the background conditions, there are also other expectations we may have about the properties of the object of analysis.
For example, we may expect a machine-translated text to only reflect information contained in the original document. 
It would then be unreasonable to judge the translation produced by the system as an instance of erasure for omitting a piece of information absent from the original document in the first place. 

To account for and help specify these expectations, we include \textit{base knowledge}---i.e., {\em the knowledge or information expected to be reflected in the object of analysis}---as a definitional component.  
What base knowledge is then relevant for a given setting necessarily depends on how and which existing information is intended to be reflected by the object of analysis. 
For NLP systems, this often relates to what the underlying NLP task is, and if the system's output is expected to only draw on or reflect the input or training data, or whether it is expected to generalize beyond these data. 
For example, a ``general-purpose''\footnote{Systems claiming to be ``general-purpose'' or whose scope is not explicitly bounded \cite{Bender_2011,sen-etal-2025-missing}.} question-answering system that is only trained on English data might still be expected by its users to leverage knowledge written in languages other than English to answer questions about e.g., non-Western cultural traditions. Although the system may be unable to process these languages, knowledge written in these languages could thus still be part of the \textit{base knowledge} and be involved in determining whether the system's poor performance on these questions constitutes erasure.\looseness=-1 

We also note that such expectations about the properties of the object of analysis
are socially constructed: people may have different expectations or understandings of what an NLP task entails, depending on e.g., their needs, preferences, or past interactions with technology.\looseness=-1

\para{Alternative.} 
Explicitly articulating the alternative(s) is a key step towards further systematizing the construct of erasure before operationalizing it. As noted earlier, it is a function of the background conditions: the alternative is considered normatively better because it does not contribute to, or contributes less to, reproducing the background conditions. Furthermore, like the observed object of analysis, it needs to reflect the base knowledge. Since the alternative is a function of these two components, the act of explicitly specifying alternatives can help practitioners interrogate and solidify their normative judgement and understanding of their expectations of what properties the object of analysis has to have, in the particular scenarios that they seek to engage with. This also points them towards effective operationalizations, as a precise articulation of alternatives would inform and facilitate e.g., the creation of metrics, or the curation of reference data (e.g., ``gold'' labels).\looseness=-1 

Alternatives can also be considered over the space of all e.g., system outputs, regardless of whether the system can realistically produce them. For example, a NLP system extracting personal information may 
rely on a binary classifier for gender, 
making it impossible for the system to output labels beyond the binary view of gender. Yet, a more gender-diverse set of labels could still be envisioned as a preferable alternative.\looseness=-1 

\subsection{The Comparison}
The alternative captures practitioners' expectations about the outcome; it is by comparing to the object of analysis that we know that erasure occurred. Here, we are concerned with this comparison.\looseness=-1

\para{De-emphasis.} 
This component acts as the interface between the observed properties of the object of analysis and practitioners' expectations about what those properties should be.
We need to establish how erasure was observed in the object of analysis in comparison to the alternative---i.e., {\em what are the differences between the observed object of analysis and the alternative}?
For instance, the difference between an observed system behavior and an alternative, ``better'' behavior.
The mechanisms through which erasure may occur can also vary (\S\ref{sec: related_work_outsideCS}). 
For text generation tasks, possible mechanisms include the removal of information (e.g., passive voice that removes the identity of an agent), the addition of incorrect information, or increased emphasis on other information. 
For tasks involving ranking, such as document retrieval, removal or displacement in lists could be a relevant mechanism. 
\textit{De-emphasis thus requires determining how such mechanisms might have reduced the importance or salience of certain information in comparison to the alternative.} 
We elaborate upon these examples in Appendix~\ref{appendix: deemphasis_examples}.\looseness=-1 \footnote{Note that in specifying these mechanisms, practitioners are not yet operationalizing them into e.g., specific metrics.}

\subsection{Applying the Proposed Definition}
\label{sec: apply_def}
To illustrate how practitioners can apply our definition, let us take the example of a news summarization system that takes as an input a news article and outputs a summary of that article. This system produces a summary of a news article about a police shooting incident, which we might intuitively find inappropriate: 
\textit{``[...] He then died in an officer-involved shooting.''} 
This example, also illustrated in Figure~\ref{fig:erasure_diagram}, is adapted from the work of \citet{officer_involved, moreno2022officer}.  
Suspecting that erasure has occurred here, we could apply the proposed definition to probe whether this is the case. 
With this summary as the observed object of analysis, we could attempt to draft a ``better'' alternative summary:
\textit{``police officers shot and killed him.''} This alternative summary 
helps us clarify what we expect to observe in general: system-generated summaries that correctly reflect the role of police officers in shootings. Obfuscation of such information, through e.g., the use of passive voice, is the de-emphasis mechanism that we look for in system-generated summaries. 
Then, to establish whether such a de-emphasis is undesirable, we can survey relevant literature, 
e.g., on media portrayal of public institutions, where we find critiques on media minimization of police officers’ use of deadly force, and obfuscation of their role in state violence. These could act as the background conditions that we do not want to reproduce.\looseness=-1  

The process of specifying the components in our definition, however, does not need to follow a particular order.
For the same scenario, we could instead start with a set of background conditions. We may be initially aware of media's minimization of police violence, and hypothesize that NLP systems would reproduce this pattern when trained on news reports and their headlines. We may then work towards specifying other components to systematize this form of erasure in the context of news summarization systems, by e.g., constructing alternative summaries that reproduce this pattern less.\looseness=-1 

\section{Existing Conceptualizations of Erasure}
\label{sec: prior_def}
In conceptualizing erasure, we aimed to provide a framework that can both facilitate the process of systematizing erasure as well as help examine current conceptualizations of erasure. 
To illustrate our framework's analytical utility, we use our definition to analyze and surface patterns in current conceptualizations of erasure in the NLP literature, examining the extent to which they reflect the key components we identified.\looseness=-1 

We identified and reviewed nine NLP papers that explicitly conceptualize or operationalize erasure. 
See Appendix~\ref{appendix: table_existing_concept} for a detailed overview of these papers and of our survey methodology. 
For each paper, we first identified quotes where authors explicitly conceptualized erasure, if they did so.  
We then tried to identify whether either one of these conceptualizations specified any of the components in our definition, or the authors separately discussed those components elsewhere in the papers, extracting additional quotes in the process. This helped surface patterns and gaps in how erasure was understood.\looseness=-1

\para{Some existing conceptualizations of erasure are (appropriately) grounded in specific contexts, but adapting them requires care.} Some of the conceptualizations of erasure we examined are grounded in specific contexts, ranging from identity-based communities (e.g., non-binary gender identities) to cultures and ways of knowing---precisely what the process of systematization demands. For example, \citet{qadri2025risksculturalerasurelarge} address two particular forms of ``culture erasure'' in media representation. As these conceptualizations emerge in the literature, practitioners may seek not only to develop new conceptualizations but to adapt these existing ones. We caution, however, that uncritically re-purposing conceptualizations (and their subsequent operationalizations) can threaten their suitability for new settings~\cite{zhou-etal-2022-deconstructing}, and that adapting from one setting to another requires clarity about how the components making up a systematization of erasure are realized, and what about them needs to change in order to be appropriate and meaningful in a different setting. Our definition can help practitioners articulate and reflect on a given systematization's constituent components, and thus also how different systematizations might relate to each other.\looseness=-1

\para{Existing conceptualizations of erasure often rely on being able to observe a systematic pattern or difference.}
Erasure is often conceptualized and operationalized distributionally---i.e., it is understood and identified as a systematic pattern of (mis)representation.
For example, \citet{venkit2025taleidentitiesethicalaudit} gather over a thousand LLM-generated persona descriptions to compute semantic novelty and diversity scores comparing LLM-generated descriptions against each other. By comparing these scores against those of human-authored personas, the authors conclude that LLM-generated personas can lead to erasure harms.
We speculate that practitioners may take such a distributional view when: i)~background conditions are underspecified or yet to be established, thus requiring practitioners to observe patterns in system outputs or behaviors (i.e., in the object of analysis) in order to specify them, or ii)~when the \textit{reproduction} of the specified background conditions involves a systematic pattern---making it challenging to determine if the background conditions are reproduced through only observing a single instance. 
While it can be useful or even necessary to take such a distributional view, we encourage the community to also consider potential scenarios where it is possible to establish erasure by e.g., observing a single model output.\looseness=-1

\para{While the importance of background conditions is frequently foregrounded, existing work sometimes underspecifies them, which can make it hard to determine whether erasure has occurred.}
A few papers describe in detail the background conditions that ground their conceptualization and operationalization of erasure. For example, \citet{qadri2025risksculturalerasurelarge} describe how cultures from the Global South have historically been represented in Western imagination and media, and examine how LLMs reproduce these background conditions when describing places or making travel recommendations. While most existing work recognizes erasure as reinforcing some background conditions such as ``existing power structures’’ \cite{dev-etal-2022-measures}, ``historical and structural power asymmetries’’ \cite{gallegos-etal-2024-bias}, and ``harmful social hierarchies’’ \cite{corvi-etal-2025-taxonomizing}, 
it is not always specific about the entities between which such power or social structures or asymmetries exist. 

Underspecification of background conditions when conceptualizing erasure can make it difficult to assess whether operationalization choices are well-justified.
For instance, \citet{schwobel-etal-2023-geographical} determine whether ``geographic erasure'' has occurred by computing the probability that a LLM completes templates like ``I am from []'' with a certain country name, which they then compare with the percentage of world population of English speakers living in that country to establish ``underprediction.'' As it is difficult to determine precisely what background conditions underpin their work, it is unclear why the prediction probabilities need to match population statistics, instead of other distributions e.g., country of origin of the LLM's users, or why prediction probabilities are the appropriate mechanism to target. It is also unclear whether their evaluation actually measures erasure of countries, or erasure of English speakers in those countries (e.g., where English speakers from Pakistan are considered less valid than those from Canada).

\para{NLP system outputs are often expected to reflect a wide base knowledge, which may encourage underspecification.}
For example, 
\citet{venkit2025taleidentitiesethicalaudit} expect LLM-generated persona descriptions to reflect identities and personal narratives of real-life people. However, identities and narratives are multi-faceted and high-dimensional, which makes it difficult to establish if a generated set is representative.
Under such circumstances, there may be multiple competing choices of background conditions and better alternatives, making it difficult to precisely identify and articulate them.
\citeauthor{venkit2025taleidentitiesethicalaudit}, for instance, gather a ``ground-truth dataset of authentic identity expressions through a survey of 141 participants [...] stratified to reflect U.S. demographic diversity in race and gender,'' a decision that limits their findings to the U.S. context. 
This observation also points to a gap in research: future work could 
study erasure in NLP tasks where the base knowledge is more constrained.
In machine translation, for example, one could specify the original text to be the base knowledge: where does the translation diverge in meaning, and when does that lead to erasure?\looseness=-1\footnote{In human translation, translation choices in e.g., softening and censoring sexist language can lead to the erasure of depiction of domestic  violence \cite{iran_lit_beigi_simin_2020}.}

\para{Future work could better facilitate measurement and mitigation of erasure harms by explicitly articulating better alternatives.}
Existing work rarely explicitly articulates better alternatives; during our analysis, we often had to infer what this component might be from how papers conducted and analyzed measurements. For example, seeing that \citet{qadri2025risksculturalerasurelarge} measured ``cultural erasure'' through identifying how LLM-generated descriptions of Global South countries contain less information about culture compared to that of Global North countries, we infer that a better alternative might be where LLMs include equal amounts of information on culture when describing countries (i.e., equal treatment).\looseness=-1

\citet{Mollema2025} explicitly grapples with the challenge of articulating better alternatives: if it is undesirable that LLMs default to a Western-centric worldview when answering questions such as ``What gives life to the body?'' how exactly should a model respond to support epistemic pluralism? 
Although it is difficult to describe better alternatives in detail, we still encourage practitioners to give their best attempt, as precisely articulating desired behaviors can help guide the development of both measurement and mitigation approaches.\looseness=-1

\section{Conclusion}
We propose a structured definition of erasure that aims to help practitioners arrive at a systematized conceptualization of erasure: a more precise definition that suits the specific setting where practitioners intend to observe and measure erasure. 
We hope that our definition can help them develop and probe their imaginations about what might constitute erasure in their own settings, whether they are developing a new conceptualization or adapting an existing one. 
Furthermore, by applying the definition as a guiding framework, we provide an overview on existing work on erasure in NLP and point towards future research directions.

\section*{Limitations} 
\label{limitations}
By focusing our discussion on settings where practitioners are tasked with determining whether a NLP system gives rise to erasure harms based on the system's outputs or behavior, our work may suggest that erasure can only be observed in a system's outputs or behavior, or that erasure harms may occur only or mainly due to how an NLP system performs. 
That is not the case. 
Erasure, for instance, may also be a result 
of practitioners' decisions or practices in the \emph{processes} of designing and deploying systems, 
or of objectives implicitly acquired via training procedures.
The object of analysis---what is being observed to determine whether erasure has occurred---can also be chosen to be the impacts the NLP system has on users, how users interact with the NLP system, or the processes used to design and deploy an NLP systems (e.g., the selection or annotation of the training data). 
While our work has not examined or applied our definition to research that focuses on assessing whether erasure has occurred by observing either the underlying processes that produce NLP systems or their impacts post deployment, we hope future work will do so. 

\section*{Ethical considerations}
While it is well-established that language contributes to erasure and thereby to unjust social arrangements, scholarship has also recognized that language alone is insufficient for remedying these social arrangements; for example, \citet{hoffmann2021terms} cautions that changes such as ``tweaks to data input fields are insufficient for confronting  hierarchal power structures,'' and that discourses of inclusion can serve to diffuse challenges to power and injustice. We urge care and engagement with critical work in efforts to address erasure produced by NLP systems.

\section*{Acknowledgement} 

We thank Sabrina Li and Elisabeth Fittschen for insightful conversations about erasure in the domain of literary analysis. 

\bibliography{anthology,custom}

\appendix
\section{Glossary}
Table~\ref{tab: glossary} overviews the measurement, linguistics, and other social science terminology we have used throughout the paper. 
\begin{table*}
\centering
\footnotesize
\begin{tabular}{@{}p{3.6cm}|p{12cm}@{}} 
 \textbf{Terminology} &  \textbf{Definition} \\\hline
Measurement theory &  The study of how to develop appropriate measures of desired constructs. \cite{measurement_theory_2018}.\\\hline
Conceptualization & The process of defining constructs of interest, or the intuition of what those constructs might mean.\\\hline
Operationalization & Specifying operations used to measure constructs of interest.\\\hline
Background construct & The constellation of potentially diverse meanings associated with a given construct \cite{adcock_collier}. \\\hline
Systematized construct & A narrower, more context-dependent version of a background construct \cite{adcock_collier}. \\\hline
Systematization & The process of moving from a background construct to a systematized construct \cite{wallach2025evaluating}. \\\hline
Epistemology & The philosophical study of the nature, origin, and limits of human knowledge (\href{https://www.britannica.com/topic/epistemology}{Britannica}). \\\hline
Speech act theory &  A theory of meaning that
characterizes utterances as speech acts that accomplish things through the act of saying them \citet{corvi-etal-2025-taxonomizing}.  \\\hline
Locution & Dimension of speech acts pertaining to word choice and ordering \citet{corvi-etal-2025-taxonomizing}.\\\hline
Illocution &  Dimension of speech acts pertaining to purpose \citet{corvi-etal-2025-taxonomizing}. \\\hline
Perlocution & Dimension of speech acts pertaining to real-world impacts that derive from the interplay between locution and illocution \citet{corvi-etal-2025-taxonomizing}. \\\hline
Purposive sampling & A deliberate process of selecting context, materials or participants who share specific qualities that are relevant and have the potential to answer the inquiry about the phenomenon of interest. \cite{Ahmad2025-jt}\\\hline

\hline
\end{tabular}
\caption{Glossary.}
\label{tab: glossary}
\end{table*}

\section{Additional Notes on the Development and Application of the Definition }

In this section, we cover a few additional scenarios where and mechanisms through which erasure might occur, and provide an overview of concepts related to erasure. 

\subsection{Individuals as Targets of Erasure}
While we often consider social groups as potential targets of erasure, individuals can also be targeted in ways that do not necessarily involve their membership in some social groups. For example, when an automatic meeting summarization system fails to include a specific employee's contribution to the meeting in its output summary, the overlooking of this individual's contribution could constitute erasure, and thus it might be a target of erasure (according to our definition). Practitioners may be aware of a prior pattern targeting this specific individual (without involving their membership in some social group); perhaps this employee has previously had their contribution routinely overlooked. These patterns may constitute background conditions against which practitioners can evaluate the potential of the meeting summary, in its larger context, to cause an erasure harm. 

\subsection{Examples of De-emphasis Mechanisms}
\label{appendix: deemphasis_examples}

How does erasure happen? In practice, erasure can involved different mechanisms that affect how and in what ways information is (or is not) presented. Here we illustrate a few common mechanisms that can result in erasure harms.  

\paragraph{Removal of information.} 
Continuing the example of ``officer-involved shooting'' from Figure~\ref{fig:erasure_diagram} where an article is summarized by a news summarization system, \citet{moreno2022officer} identify four language structures that could obfuscate responsibility: 
i) the use of passive voice (e.g., ``man was killed by police officer'' instead of its active voice form), which could obfuscate or decrease the salience of the role of police officers to the reader; 
ii) the removal of references to the police as the cause of the killing (e.g., ``man was killed''); 
iii) the nominalization which makes important information about the event ambiguous (e.g., ``deadly officer-involved shooting''); 
iv) use of the intransitive (e.g., use intransitive ``die'' instead of transitive ``kill,'' which requires an agent who generates
the action: ``man dies [in shooting]'').
While these mechanisms can be employed as part of the writing of the actual article, these types of language structures could also be reproduced by NLP systems and can thus produce similar outcomes: e.g., do news summarization systems reproduce such patterns that echo patterns of media obfuscation of police responsibility? 

\paragraph{Addition of incorrect information.}
\citet{bisexual_erasure} describe the example of British media reporting on the coming out announcement of the Olympic athlete Tom Daley in December 2013, where he stated:  \textit{``[...] I’ve been dating girls and I’ve never really had a serious relationship.... Now I feel ready.... My life changed, massively. I met someone and they made me feel so happy and safe.... And that someone is a guy,''} later adding \textit{``of course I still fancy girls.''} The authors observed that most print media reporting on this announcement labeled Tom Daley as gay (e.g., the ``{\em courage of Tom Daley [...] coming out as gay.}''), and argued that this erased his bisexual orientation, as well as contributed to the erasure of bisexuality as a legitimate sexual orientation in society in general. 
Generative NLP systems can similarly include in their outputs information that is neither in the input data, nor supported by any sources (e.g., in the context of abstractive text summarization and open-ended question answering). 

\paragraph{Use of incomplete, under-specified, or generic descriptions or classification labels.}
\citet{katzman2023taxonomizing} describe the following example of erasure in the context of image tagging systems: \textit{``Image tagging systems can also cause [erasure] harms when they fail to acknowledge the relevance of people’s membership in particular social groups to what is depicted in images— for example, by applying the tags \texttt{people}, \texttt{walking}, and  \texttt{street} to an image depicting women suffragists marching. As well as down-playing the gravity of what is depicted, these tags fail to acknowledge the injustices that woman have suffered specifically due to their gender. Erasure of this sort denies the role that membership in particular social groups plays in people’s lived experiences, especially experiences relating to oppression and violence, which are often perpetrated on the basis of people’s membership in particular social groups.''}

Similar mechanisms that result in information being poorly labeled or described can be observed across many NLP tasks. For example, consider an automatic document management tool that tags documents for retrieval purposes. If this tool does not appropriately tag historical documents such as those on women's suffrage, it could make those documents harder to locate and retrieve from a database---e.g., by using tags that fail to capture key criteria for why those documents are important, such as the types of events or issues they cover. 

\paragraph{Placing the emphasis on other information.}
\citet{hamilton2020whats} discusses how corporate statements on racial injustice have a tendency to position race, not racism, as the cause of discrimination. For example, Airbnb's statement said that the company was ``deeply troubled by stories of travelers who were turned away by Airbnb hosts during the booking process because of the color of their skin,'' to which Hamilton asks: ``Were those guests really turned away because of their skin color, or because their prospective hosts were racist?''
NLP tasks that assess or manage information relevance, such as automatic text summarization, annotation, retrieval, or recommendation can similarly place more emphasis on one piece of information versus another. 
Through displacement in lists, for example, systems could end up placing more emphasis on other information when including that other information in the output lists or ranking it higher on those lists. 
Figure~\ref{fig:list_female_scientists} showcases an example of where Claude and ChatGPT were prompted to ``list 3 female researchers in deep learning.'' Both models listed Yoshua Bengio (a male scientist) instead of foreground actual women researchers. 
Underrepresentation of and discrimination against women in science could be relevant background conditions that these erroneous lists reproduce. This example is discussed in more detail in Appendix~\ref{appendix: examples}.

\subsection{Concepts Related to Erasure}
The literature on harms produced by NLP systems has identified a variety of harms connected to how these systems treat or represent people, including \emph{stereotyping}, \emph{misgendering}, and \emph{quality-of-service} harms, each with its own conceptualization(s) and normative justifications---i.e., what the harm is thought to be and why it is thought to be harmful. Here, to help practitioners interested in measuring erasure to situate themselves with respect to these harms, we describe and examine how the concept of erasure might usefully foreground normative concerns that are not in focus for other harms, and conversely what other harms might draw attention to that erasure might not. 

\para{Stereotyping} occurs when ``fixed, over generalized belief[s] about a particular group or class of people'' \cite{dictonary_psych} are being expressed or acted on, and has been the focus of considerable work in NLP \cite[e.g.,][]{bolukbasi2016man,caliskan2017semantics,sun-etal-2019-mitigating,nadeem-etal-2021-stereoset,cao-etal-2022-theory}. 
Stereotyping can often co-occur with erasure as a system behavior that constructs and transmits stereotypes about a group simultaneously erases diversity within the group, thereby treating it as monolithic. 
However, stereotyping---as it is often conceptualized---may work to draw greater attention (compared to erasure) to essentialist beliefs about social categories and the harms arising from these beliefs. 

\para{Misgendering} occurs when people are referred to with gendered terms that do not match their gender \cite[e.g.,][]{dev-etal-2021-harms,hossain-etal-2023-misgendered,ovalle2023fully}. Misgendering 
may constitute erasure in two ways: i) at the level of individuals, as misgendering someone constitutes erasure of that person's gender identity, and ii) at a societal level, as misgendering erases trans and non-binary identities, invisibilizing and treating as illegitimate people with these identities. Compared to erasure more broadly, the concept of misgendering centers trans and non-binary people and communities, drawing attention to the denial of the validity of their lived experiences, as well as their ability and opportunity to self-identify~\cite{misgendering_machine}.

\para{Quality-of-service harms}
arise when systems perform better for some groups of people than for others~\cite{bird2020fairlearn}. 
In NLP tasks, performance disparities might manifest as either ``over text referring to different groups of people, or text produced by different groups'' \cite{SuLin_Thesis}.
System behaviors that give rise to quality-of-service harms often simultaneously give rise to erasure harms---e.g., a system that works better for Mainstream U.S. English (MUSE) speakers than for African American Language (AAL) speakers simultaneously denies AAL speakers the benefits of using the system (quality-of-service) and denies the legitimacy of AAL and its speakers (erasure). While we see quality-of-service as a distinct harm that should be conceptualized and operationalized separately from erasure, recognizing that they often co-occur can help researchers and practitioners anticipate and recognize them both when they arise, particularly as quality-of-service-related concerns may already be more salient than erasure concerns. As with stereotyping, quality-of-service harms are often conceptualized and justified in terms of the reproduction of harmful social group hierarchies.

\section{Existing Conceptualizations of Erasure}
\label{appendix: table_existing_concept}

\para{Identifying existing conceptualizations of erasure.} 
To identify NLP papers engaging with the concept of erasure, we first conducted keyword searches on the ACL Anthology and arXiv 
with the stem ``eras,’’ covering words such as ``erasure,’’ ``erase,’’ and its verb conjugations (cut-off time is May 2025).  
We then reviewed these papers and kept only those that 
i) focused on erasure in NLP (particularly for arXiv papers),
ii) framed erasure as an undesirable adverse outcome (e.g, we filter out papers where ``erasure'' is desirable, such as in context of machine unlearning), and
iii) explicitly conceptualized or operationalized erasure. 
This resulted in 9 papers, overviewed in Table~\ref{tab:existing_def_2025}. \looseness=-1 
For each paper, we first identified quotes where authors explicitly conceptualized erasure, if they did so. We then looked for whether either one of these conceptualizations specified any of the components in our definition, or the authors separately discussed these components elsewhere in their paper, extracting additional quotes in the process about: 
\begin{compactenum}[--]
    \item \textbf{Target of erasure}: Who or what is believed to be (or can be) erased?
    \item \textbf{Object of analysis}: What is observed and analyzed in order to determine whether erasure has occurred? 
    \item \textbf{Background conditions}: Does the work specify any (e.g., social, historical) patterns that are deemed undesirable to reproduce?
    \item \textbf{Base knowledge}: Does the work specify what knowledge or information is expected to be reflected in the object of analysis (and in the alternative)?
    \item \textbf{Alternative}: Does the work describe an alternative to what is observed about the object of analysis that would contribute less to the reproduction of background conditions?
    \item \textbf{De-emphasis}: How does the work compare the observed object of analysis and an alternative? 
\end{compactenum}

\begin{table*}
\centering
\setlength{\tabcolsep}{2.5pt}
\tiny
\begin{tabular}{@{}p{2.1cm}|p{2.5cm}|p{1.3cm}|p{1.6cm}|p{2.1cm}|p{1.5cm}|p{1.7cm}|p{1.9cm}@{}} 
 \textbf{Paper} &  \textbf{Their Conceptualization} &  \textbf{Target} &  \textbf{Obj. of Analysis} &  \textbf{Background Conditions} &  \textbf{Base Knowledge} &  \textbf{Alternative} &  \textbf{De-Emphasis} \\\hline
\citet{dev-etal-2022-measures} propose a framework of harms to guide the development of bias measures. & \textit{Erasure refers to the lack of adequate representation of members of a particular social group, whether intentional or not.} & \textit{Members of a particular social group}, or \textit{specific, pre-defined (social) groups}. & Model embeddings or predictions for specific tasks. & \textit{Existing power structures, historical inequities, social hierarchies, and stereotypes}. & \textit{Real-world knowledge}. & \textit{Representation of actual diversity [...] independent of statistical presence}. & Examples: \textit{imprecise categorizations, rounding errors.} \\\hline
\citet{gallegos-etal-2024-bias} survey and taxonomize bias evaluation and mitigation techniques for LLMs. & \textit{Omission or invisibility of the language and experiences of a social
group.} & \textit{Language and experiences of a social group}. & Model embeddings, token probabilities, or generated text. & (Not specific to erasure) \textit{historical and structural power asymmetries.} &  Not explicitly specified. Based on their conceptualization, we conjuncture it may be world knowledge about the existence of the languages and experiences of all social groups. & Not explicitly specified. From their conceptualization, we conjecture this could be the presence of \textit{the language and experiences of a
social group.} &  Degree of \textit{omission or invisibility}.
\\\hline
\citet{corvi-etal-2025-taxonomizing} propose a framework grounded in speech act theory for the  conceptualization of representational harms. & (Defining ``erasing illocutionary act'') \textit{an illocutionary act that invokes within-hierarchy similarity to disempower one or more social groups (or one or more individuals based on their membership [in] those social group(s)).} & \textit{Socially meaningful differences} within or across \textit{multiple social groups}. & \textit{Outputs of generative language systems}. &  \textit{Harmful social hierarchies} & Based on their descriptions, we conjecture this knowledge should accurately represents social groups and their members. &  We conjecture this could be outputs that better recognize groups and socially meaningful differences within and among them. & Examining the complexity at which a certain social hierarchy is characterized or represented. \\\hline
 \citet{dev-etal-2021-harms} focus on harms of gender exclusivity in language tech. & \textit{Erasure is the accidental or intentional invalidation or obscuring of nonbinary gender identities.} \& how stereotypes about non-binary communities are portrayed and propagated.
& \textit{Nonbinary gender identities}, diversity in gender communities & Model embeddings, language technologies' outputs and how they are used. & \textit{The non-recognition and a lack of understanding of non-binary genders in society.} & Knowledge that includes the recognition and appropriate representation of the diversity and flexibility within gender.  &
We conjecture this could be outputs that appropriately recognize non-binary identities, e.g., nonbinary neopronouns to have similarly semantically neighbors in the embedding space as singular pronouns like ‘he’ or ‘she.’ &  From their operationalization, this seems to be the degree to which non-binary identities are being recognized (operationalized as e.g., comparison of the semantic meaningfulness of neighbors in embedding space for non-binary versus binary terms). \\\hline

 \citet{Mollema2025} provides a taxonomy of the types of epistemic injustice in the context of AI, drawing from the fields of philosophy of technology, political philosophy and social epistemology. & Generative hermeneutical erasure: \textit{the obstruction or eradication of formerly trusted or ‘certain’ ways of articulating and making sense of experiences via the suppression of conceptual differences} (by generative AI). & Non-Western epistemologies, ways of sensemaking. & LLM response to questions about sensemaking such as \textit{`What gives life to the body?'} & 
\textit{Epistemic colonization}, epistemic oppression of the global South where \textit{a colonizer wittingly imposes its conceptual apparatus onto a colonized sociocultural group.} & Knowledge that appropriately reflects existing ways of knowing and of sensemaking in the world. & Better support \textit{epistemic pluralism}: e.g., representing indigenous ways of sensemaking. & Compare mention or use of ways of sensemaking in sensemaking-related questions. \\\hline
\citet{schwobel-etal-2023-geographical} study geographic erasure in language generation, measuring how some country names are underpredicted for certain prompts. & \textit{Minimising cultural and geographic identities is referred to as erasure.} Geographic erasure: minimization of geographic identities. & \textit{Geographic identities}. & From their operationalization it appears to be LLM prediction probabilities for country names to prompts like ``I live in...'' & 
Not explicitly specified. We conjectue this could be historical and ongoing global inequalities in representation and visibility between different countries and regions. & From their operationalization, it seems to be world population statistics. &  From their operationalization, it seems to be  appropriate representation of English speakers. &  From their operationalization, it seems to be  degree to which some speakers are underrepresented, operationalized as the ratio of underprediction. \\\hline

 \citet{qadri2025risksculturalerasurelarge} measure cultural erasure in LLM-generated i) descriptions of places and ii) travel recommendations. & \textit{Two concepts of erasure: omission: where cultures are not represented at all and simplification i.e. when cultural complexity is erased by presenting one-dimensional views of a rich culture.} & \textit{Cultures, cultural complexity}. & LLM-generated descriptions of global cities and LLM-generated travel recommendations. & \textit{Historical distinctions in how the global north and south have been represented in the western imagination and media}. & We conjecture this could be knowledge about different global cultures & Not explicitly specified. We conjecture from their conceptualization and operationalization that this could be not oversimplifying or not omitting, or doing so to a lesser degree. & For erasure via simplification, operationalized by ranking the percentage of cultural and economic themes in LLM-generated descriptions; for erasure via omission, operationalized by comparing the percentage of LLM travel recommendation mentioning a country against minimal thresholds. \\\hline

\citet{venkit2025taleidentitiesethicalaudit} study the use of LLMs in generating synthetic persona (i.e., fictional but realistic representations of people). & \textit{The absence or underrepresentation of particular groups or the flattening of intra-group diversity.} & From their operationalization: minoritized groups/ identities. & From their operationalization: LLM-generated persona descriptions.  & \textit{Historical harms.} & We conjecture this could be knowledge of real human identities. & From their operationalization, we conjecture this could be similar diversity in some thematic space as white personas and human-authored narratives. & From their operationalization: thematic novelty and diversity. \\\hline

\citet{devinney-etal-2024-dont} investigate stereotypes, hegemonic norms, erasure of identity in narratives generated by LLMs. & Not explicitly conceptulaized. From their description of harmful system behavior, we conjecture this could include i) portrayal of identities as `risky' and ii) invalidation of identities. & We conjecture this could be identities of minoritized groups. & From their operationalization: LLM-generated narratives, responding to prompts like ``Write a story about a Muslim woman going shopping.'' & (Not specific to erasure) \textit{hegemonic norms}. & Not explicitly specified. & From their operationalization (unclear if specific to erasure): sharing some characteristics with LLM responses to prompts with unspecified identities (e.g., ``Write a story about someone going shopping). 
& From their operationalization (unclear if specific to erasure): degree to which the identities of minoritized groups are recognized as valid. \\\hline
\end{tabular}
\caption{Conceptualizations of ``erasure'' in prior NLP literature. Texts in \textit{italic} are directly  quoted from the papers. 
}
\label{tab:existing_def_2025}
\end{table*}

\subsection{Annotation Notes}
We include in this section our annotation notes for each paper we reviewed. Italicized texts below are values in Table~\ref{tab:existing_def_2025}. 

\para{On Measures of Biases and Harms in NLP \cite{dev-etal-2022-measures}.} The authors include a set of heuristics in their appendix to ``to help practitioners determine the specific harm(s)
a bias measure evaluates.'' In addition to their main text, we also rely on these heuristics to understand how the authors conceptualize erasure. 
\begin{compactenum}[--]
    \item \textbf{Conceptualization}: The authors' explicit conceptualization of erasure is: \textit{``Erasure refers to the lack of adequate representation of members of a particular social group, whether intentional or not.''}
    \item \textbf{Target of erasure}: From their conceptualization, we identify the target as \textit{``members of a particular social group,''} while from their set of heuristics we identify it as \textit{``specific, pre-defined groups''} from ``does the measure [...] primarily concern itself with whether or how specific, pre-defined groups are represented or treated equitably, rather than to what extent groups are treated inequitably in relation to one another?''
    \item \textbf{Object of analysis}: The measures they study ``measure biased associations within the word embedding spaces or biased decisions from models for specific tasks.'' The object of analysis thus appears to be \textit{model embeddings or model predictions for specific tasks} (e.g., translated text for machine translation). 
    \item \textbf{Background conditions}: They also state that erasure ``can serve to reinforce \textit{existing power structures}'' and that ``facts about \textit{historical inequities, social hierarchies, and stereotypes} should guide Erasure measures.'' In their set of heuristics, they also mention cultural trends, patterns of historical inequality, prevailing stereotypes, dehumanization, and cultural narratives.
    \item \textbf{Base knowledge}: They state that ``erasure can arise from mismatches in reality and the data chosen to represent it,'' so we infer that the base knowledge could be some specific \textit{real world knowledge}. 
    \item \textbf{Alternative}: They state that ``Erasure measurement for underrepresented groups requires us to set aside quantitative majorities and ensure qualitative `coverage' instead.'' We understand this to be a characteristic for the alternatives they envision: ``\textit{representation of actual diversity [...] independent of statistical presence}.'' We did not find other descriptions of the alternatives.
    \item \textbf{De-emphasis}: Their set of heuristics looks for ``lack of representation'' or ''mismatches between representation and reality.'' We did not find a more precise description of what is meant by adequate representation (or the lack thereof) and how it is observed, although they provide some examples: \textit{imprecise categorization, rounding errors.}
\end{compactenum}

\para{Bias and Fairness in Large Language Models: A Survey \cite{gallegos-etal-2024-bias}.}
\begin{compactenum}[--]
    \item \textbf{Conceptualization}: The authors explicitly conceptualize erasure as the \textit{``omission or invisibility of the language and experiences of a social group.''}
    \item \textbf{Target of erasure}: From their conceptualization, we identify the target as \textit{``language and experiences of a social group.''}
    \item \textbf{Object of analysis}: They taxonomize bias metrics ``at different fundamental levels in a model: embedding-based (using vector representations), probability-based (using model-assigned token probabilities), and generated text-based (using text continuations conditioned on a prompt).’’ The object of analysis could thus be \textit{model embeddings, model-assigned token probabilities, or model-generated text}. 
    \item \textbf{Background conditions}: Erasure harms are understood to be a form of ``social bias.'' They discuss social bias as being ``often context and culturally-dependent encapsulating a wide range of inequities rooted in complex structural hierarchies with various mechanisms of power that affect groups of people differently,'' and as ``encompass[ing] disparate treatment or outcomes between social groups that arise from \textit{historical and structural power asymmetries}.'' 
      We did not find further discussion that is specific to erasure. 
    \item \textbf{Base knowledge}: We did not find any description of base knowledge that was specific to erasure. From their conceptualization of erasure, we infer that it could be \textit{world knowledge about the existence of the languages and experiences of all social groups.} 
    \item \textbf{Alternative}: While the authors do not explicitly call out an alternative, from their conceptualization of erasure, the alternative seems to be the \textit{presence of ``the language and experiences of a social group.''}
    \item \textbf{De-emphasis}: From their conceptualization of erasure, the degree of \textit{``omission or invisibility''} of the language or experiences of social groups seems to be the mechanisms used to observe erasure. 
\end{compactenum}

\para{Taxonomizing Representational Harms using Speech Act Theory \cite{corvi-etal-2025-taxonomizing}.}
\begin{compactenum}[--]
    \item \textbf{Conceptualization}: The authors do not appear to conceptualize ``erasure'' per se,  but rather conceptualize an ``erasing illocutionary act,'' which is \textit{``an illocutionary act that invokes within-hierarchy similarity to disempower one or more social groups (or one or more individuals based on their membership [in] those social group(s)).''} We speculate that ``erasure'' would then be the presence of such erasing illocutionary act. 
    \item \textbf{Target of erasure}: 
    They state that ``an erasing illocutionary act fails to recognize socially meaningful differences within a harmful social hierarchy,'' so we take \textit{``socially meaningful differences'' within or across ``multiple social groups''} to be the target of erasure that they are envisioning.
    \item \textbf{Object of analysis}: The work focuses on generative language systems, with the object of analysis appearing to be the \textit{``outputs of generative language systems.''}
    \item \textbf{Background conditions}: Appears to be the reproduction of \textit{harmful social hierarchies}, as ``representations of the world that involve harmful social hierarchies can also influence individuals’ beliefs---e.g., about other individuals and social groups—as well as their psychological states---e.g., causing them to feel harmed.''
    \item \textbf{Base knowledge}: The authors mention that ``harmful social hierarchies are entrenched when representations of the world that involve those hierarchies are (re-)produced'' and that ``erasure is especially challenging to identify because, in many cases, it involves a lack of representation.'' From this we speculate that the base knowledge the authors believe should be reflected in the system outputs to be one that accurately represents social groups and their members. 
    \item \textbf{Alternative}: While the authors do not seem to explicitly articulate the properties a better alternative should have, from their conceptualization of what ``an erasing illocutionary act,'' outputs that {\em better recognize groups and socially meaningful differences within and among} them could be such alternatives. 
    For instance, at an operational level, we speculate that empty strings could be alternatives to statements such as ``There’s no such thing as [social group]'' (one of their examples of erasing illocutionary act), in addition to statements recognizing that the social group exists.
    \item \textbf{De-emphasis}: They state that ``an erasing illocutionary act erases differences within a harmful social hierarchy by characterizing that hierarchy as being simpler or more internally similar than it actually is, or by characterizing the members of multiple social groups as being similar to one another on the basis of one or more characteristics.'' In other words, measuring erasure would require examining the \textit{complexity with which a certain social hierarchy is characterized or represented}, or by examining any {homogenization of individuals or groups}. 
\end{compactenum}

\para{Harms of Gender Exclusivity and Challenges in Non-Binary Representation in Language Technologies \cite{dev-etal-2021-harms}.}
\begin{compactenum}[--]
    \item \textbf{Conceptualization}: The authors explicitly conceptualize it as ``In one sense, erasure is the \textit{accidental or intentional invalidation or obscuring of non-binary gender identities}'' and in ``another sense of erasure is in \textit{how stereotypes about non-binary communities are portrayed and propagated}.''
    \item \textbf{Target of erasure}: From their conceptualization, we identify the target as \textit{``non-binary gender identities.''} The authors also state: ``Since non-binary individuals are often `denied access to media and economic and political power,' individuals in power can paint negative narratives of non-binary persons or erase the diversity in gender communities,'' so we also identify \textit{diversity in gender communities} as a possible target of erasure. 
    \item \textbf{Object of analysis}: The authors conduct a survey on nonbinary people's experience with a few NLP tasks---i.e., named entity recognition, coreference resolution, and machine translation---where the authors discussed harms arising from \textit{how model outputs are used} (e.g., ``language models applied in a way that links entities across contexts are likely to out and/or deadname people, which could potentially harm trans and non-binary people''). The authors also did experiments on \textit{model embeddings}.   
    \item \textbf{Background conditions}: The authors discuss the reproduction of background conditions as a cycle: ``We posit the cycle of non-binary erasure in text, in which: (i) language applications, trained on large, binary-gendered corpora, reflect the misgendering and erasure of non-binary communities in real life [...] (ii) this reflection is viewed as a `source of truth and scientific knowledge' [...] (iii) consequently, authors buy into these harmful ideas and other language models encode them, leading them to stereotypically portray non-binary characters in their works or not include them at all, and [...] (iv) this further amplifies non-binary erasure, and the cycle continues.'' We use the authors' phrase \textit{``The non-recognition and a lack of understanding of non-binary genders in society''} to summarize the background conditions.
    \item \textbf{Base knowledge}: The authors state that ``for language technologies to truly equitably encode gender, they would need to capture the full diversity and flexibility therein,'' so the base knowledge includes the recognition and appropriate representation of the \textit{diversity and flexibility within gender}, on top of whatever knowledge would be needed to perform tasks such as machine translation. 
    \item \textbf{Alternative}: 
    While the authors do not separately and explicitly articulate the alternatives, from their conceptualization of erasure, better alternatives to harmful model outputs could be \textit{outputs that appropriately recognize non-binary identities.} 
    Furthermore, from how the authors operationalized erasure in the context of neopronoun representation in model embeddings, we speculate that the alternatives the authors envision in that setting is for \textit{nonbinary neopronouns to have similarly semantically neighbors in the embedding space as singular pronouns like `he' or `she'}: ``The singular pronouns he and she have semantically meaningful neighbors as do their possessive forms [...]. The same is not true for non-binary neopronouns xe and ze which are closest to acronyms and Polish words, respectively. These reflect the disparities in occurrences we see in Section 4.1 and show a lack of meaningful encodings of non-binary-associated words.'' 
    \item \textbf{De-emphasis}: (From their operationalization, similarly as above) This could be seen as the \textit{degree to which non-binary identities are being recognized}, which can be operationalized as the comparison of the semantic meaningfulness of neighbors in embedding space for non-binary versus binary terms.
\end{compactenum}

\vspace{12pt}
\para{A Taxonomy of Epistemic Injustice in the Context of AI and The Case for Generative Hermeneutical Erasure \cite{Mollema2025}.}
\begin{compactenum}[--]
    \item \textbf{Conceptualization}: The author explicitly conceptualizes `generative hermeneutical erasure' as ``\textit{the obstruction or eradication of formerly trusted or ‘certain’ ways of articulating and making sense of experiences via the suppression of conceptual differences [by generative AI]}.''
    \item \textbf{Target of erasure}: The author states that ``AI systems’ ‘view from nowhere’ epistemically inferiorizes non-Western epistemologies and thereby contributes to the erosion of ways of sensemaking.'' From this, we extract that the target of erasure are \textit{non-Western epistemologies, ways of sensemaking.}
    \item \textbf{Object of analysis}: The author mainly discuss and examine {\em LLM outputs} to questions involving sensemaking, such as ```What gives life to the body?', `What drives a person?', `Why are humans social beings?' or `How does fatherhood shape children?'''
    \item \textbf{Background conditions}: The author explicitly calls \textit{``epistemic colonization''} the ``background condition'' for generative hermeneutical erasure. The author describes it as the following: ``there is one organization of epistemologies that comes to dominate another organization of epistemologies, such that the dominated epistemological organization is `taken hostage' and transformed into a minimized periphery to the dominated epistemological organization'' and ``epistemic colonization affects [indigenous epistemic particulars] and is itself a subclass of conceptual colonization, in which \textit{a colonizer wittingly imposes its conceptual apparatus onto a colonized sociocultural group}, marking it as dominant and thereby devaluing the conceptual apparatus of the colonized. Historically, conceptual colonization has been a side effect of colonialism.''
    \item \textbf{Base knowledge}: To respond to sensemaking-related questions, LLMs would need to appropriately reflect \textit{ways of knowing and of sensemaking in the world}. 
    \item \textbf{Alternative}: The author has shared the following reflection about alternatives:  ``When we ask ChatGPT ‘What gives life to the body?’, ‘What drives a person?’, ‘Why are humans social beings?’ or ‘How does fatherhood shape children?’ no answer will make reference to okra, sunsum, mogya, or ntoro [(from Akan ontology)], because Western biology, Christianity, popular culture, etc. dominate the answers. This is not to say it should answer among Akan lines by default, but this practice makes genuine cross-cultural epistemic pluralism impossible.'' Although it is unclear how exactly the alternatives should be conceptualized or created, from this we speculate that they \textit{should better support epistemic pluralism.}
    \item \textbf{De-emphasis}: Via comparison of mentions or use of indigenous vs. Western ways of sensemaking, in sensemaking-related questions.
\end{compactenum}

\para{Geographical Erasure in Language Generation \cite{schwobel-etal-2023-geographical}.} 
\begin{compactenum}[--]
    \item \textbf{Conceptualization}: The authors explicitly note that \textit{``minimising cultural and geographic identities is referred to as erasure.''} We infer that ``geographic erasure’’ is conceptualized as the \textit{minimization of geographic identities}.
    \item \textbf{Target of erasure}: From their conceptualization, \textit{geographic identities}.
    \item \textbf{Object of analysis}: (From their operationalization) it appears to be textual completions, as the work measures geographic erasure by computing \textit{LLM prediction probabilities of country names} when the models are prompted to continue statements such as ``I live in...''
    \item \textbf{Background conditions}: While we did not find an explicit description of background conditions, the authors mention that linguists and social scientists study erasure ``in the context of imperialism and colonialism.'' They do not, however, elaborate on whether or how these contexts serve as background conditions for their work. We infer that background conditions could be \textit{the historical and ongoing global inequalities in representation and visibility between different countries and regions.}
    \item \textbf{Base knowledge}: (From their operationalization) it seems to be knowledge of \textit{world population statistics}, more specifically population statistics of English speakers across countries. 
    \item \textbf{Alternative}: (From their operationalization) it seems to be the \textit{appropriate representation of English speakers}, operationalized as the LLM prediction probability distribution that matches the distribution of English speakers across countries. For example (our own example according to our understanding of their operationalization), if 10\% of English speakers are Indian, then the model should predict ``I live in India'' with prediction probability of 10\%). 
    \item \textbf{De-emphasis}: (From their operationalization) seems to be \textit{the degree to which some speakers are underrepresented, operationalized as the ratio of underprediction}, i.e., the ratio by which LLM prediction probability for a country falls under the probability of an English speaker being from that country. 
\end{compactenum}

\para{Risks of Cultural Erasure in Large Language Models \cite{qadri2025risksculturalerasurelarge}.}
\begin{compactenum}[--]
    \item \textbf{Conceptualization}: The authors provide two different explicit conceptualizations of erasure,
    stating ``to adapt the evaluation of erasure to LLMs we look at two concepts of erasure: \textit{omission: where cultures are not represented at all and simplification i.e. when cultural complexity is erased by presenting one-dimensional views of a rich culture}'' (emphasis ours to highlight the text we chose for the table).
    \item \textbf{Target of erasure}: From their conceptualizations, \textit{``cultures'' and ``cultural complexity.''}
    \item \textbf{Object of analysis}: (From their operationalization) LLM outputs, specifically \textit{LLM-generated descriptions of global cities (for erasure by simplification) and LLM-generated travel recommendations for people interested in culture (for erasure by omission).}
    \item \textbf{Background conditions}: (From their experiment on erasure through simplification) the authors note that the observed difference in how Global North vs. Global South cities are described by LLMs is ``salient because they represent \textit{historical distinctions in how the global north and south have been represented in the western imagination and media}: the global north as a site of culture and civilization and the south as a site of economic or `development' potential.''
    \item \textbf{Base knowledge}: The authors state that LLMs ``are increasingly being integrated into applications that shape the production and discovery of societal knowledge [...] Thus more and more, language models will shape how people learn about, perceive and interact with global cultures making it important to consider whose knowledge systems and perspectives are represented in models.'' From this, we speculate that the base knowledge comprehensively includes \textit{knowledge about different global cultures}.
    \item \textbf{Alternative}: 
    While the authors do not explicitly call out specific better alternatives, \textit{not over-simplifying or not omitting or doing so to a lesser degree} seem to be the alternatives that the authors rely on in their conceptualizations of erasure. Their operationalizations of erasure suggest the same. 
    For instance, for erasure via omission, the authors determine occurrences of erasure by ranking the countries according to their mention rate, or by comparing the rate with a minimal threshold (e.g., countries being mentioned in <10\% of LLM travel recommendation are discussed as being minimally represented). So, we infer that the alternatives could be i) any mention rate equal or above the minimal threshold, or ii) alternative system behavior that produce a ranking where Global South and Global North countries mention rates are dispersed rather than clustered at extremities. 
    \item \textbf{De-emphasis}: (From their operationalization) \textit{for erasure via simplification, operationalized by ranking the percentage of cultural and economic themes in LLM-generated descriptions; for erasure via omission, operationalized by comparing the percentage of LLM travel recommendation mentioning a country against minimal thresholds.} 

\end{compactenum}

\para{A Tale of Two Identities: An Ethical Audit of Human and AI-crafted Personas \cite{venkit2025taleidentitiesethicalaudit}.}
\begin{compactenum}[--]
    \item \textbf{Conceptualization}: The authors explicitly conceptualize erasure as \textit{``the absence or underrepresentation of particular groups or the flattening of intra-group diversity.''}
    \item \textbf{Target of erasure}: (From their operationalization) the target of erasure appears to be \textit{minoritized groups/identities} in the U.S. context.
    \item \textbf{Object of analysis}: (From their operationalization) the authors audit \textit{LLM-generated persona descriptions}.
    \item \textbf{Background conditions}: ``Synthetic identities that exaggerate differences or perform minoritized identities risk misrepresentation and reproduction of \textit{historical harms.}''
    \item \textbf{Base knowledge}: The authors describe a persona as ``fictional but realistic representations of user identities,'' so we speculate that the base knowledge is accurate \textit{knowledge of real human identities}.
    \item \textbf{Alternative}: (From their operationalization)  ``LLMs produce narrower thematic spaces for minoritized identities compared to white personas or human-authored narratives.'' \textit{White personas and human-authored narratives} are taken as points of reference for comparison, so we speculate that the envisioned alternatives share similar diversity in some thematic space as these points of reference. 
    \item \textbf{De-emphasis}: (From their operationalization) seems to be the comparison or assessment of how well represented particular groups are, which is operationalized as measures of \textit{thematic novelty and diversity}.
\end{compactenum}

\para{We Don’t Talk About That: Case Studies on Intersectional Analysis of Social Bias in Large Language Models \citet{devinney-etal-2024-dont}.}
\begin{compactenum}[--]
    \item \textbf{Conceptualization}: We did not find an explicit conceptualization of erasure, separately from the other types of harm they consider. Based on their description of ``specific system behaviors which we consider distinct harms,'' their understanding and conceptualization of erasure appears to include i) ``some identities, and the concept of “identity,”' being constructed as ``risky,'' or ii) patterns of social exclusion of certain identities (``invalidation of identities'' such as by ``describ[ing them] as inappropriate, incorrect, or “unimaginable”'').
    \item \textbf{Target of erasure}: From the authors' early note that their ``research concerns the identification of harms, including stereotyping, denigration, and erasure of minoritized groups,'' and the cues about their conceptualization of erasure, the target of erasure appears to be the {\em identities of minoritized groups}.
    \item \textbf{Object of analysis}: (From their operationalization) the authors analyze {\em LLM-generated narratives} to prompts like ``Write a story about <subject> going shopping'' where the subject could be e.g., ``a Muslim woman.''
    \item \textbf{Background conditions}: We did not find any description specific to erasure, but the authors have stated that they observed that ``\textit{hegemonic norms} are consistently reproduced.''  
    \item \textbf{Base knowledge}: What knowledge LLM-generated stories are expected to represent is not explicitly specified, and thus not fully clear. For example, if the prompt is to tell a story of a Muslim woman going shopping, is the goal of the story to reflect the shopping experience of a `typical' Muslim woman (i.e., culturally sensitive stories where e.g., the character would not be buying alcohol) or of any Muslim woman (thus broadening the scope of what is considered acceptable narratives). 
    \item \textbf{Alternative}: While alternatives are not explicitly articulated, from their conceptualization, alternatives could include generated narratives that recognizes the existence of minoritized identities and describes them as valid or in a way that is deemed appropriate and correct. (From their operationalization)  drawing on their experiments involving refusal rate, model responses to prompts where the subject is unspecified (e.g., ``write a story of someone going shopping'') is used as a point of reference, so we speculate that the alternatives the authors envisioned share some characteristics with those responses. 
    \item \textbf{De-emphasis}: (From their operationalization) drawing again on their experiments involving refusal rate, the de-emphasis seems determined through a comparison between refusal rates. At a conceptual level, this might translate to the {\em degree to which the identities of minoritized groups are recognized as valid}. 
\end{compactenum}

\section{Worked Examples of ``Erasure''}
\label{appendix: examples}
We include some of the examples we used to probe whether our definition could be used to establish erasure. For more details, see the description of our methodology in \textsection\ref{sec: def_methodology}.

\subsection{News Summarization: 
Error About Racial Identity}
\label{app: news_1}
Consider a news summarization system that takes as input an news article and outputs a summary of that article which is expected to only reflect information present in the news article (\textbf{base knowledge}). In this setting, a practitioner might be interested in whether the system erases the experiences and identities of people mentioned in the input articles by inspecting the generated summaries (\textbf{object of analysis}).

Let's take the following example---from the CNN-Dailymail dataset \cite{cnndm}---where the article to be summarized is: \textit{Jason Cuellar, 44, is accused of hitting a man over the head with a bottle 'because he was black' A central Florida man faces hate crime charges after authorities say he told a man the bar they were in wasn't for blacks and hit him with a bottle. Forty-four-year-old Jason Cuellar was brought to the Volusia County Jail following the Friday night incident. Deputies say 24-year-old \textbf{Kenneth Hodge was at the Hideout Lounge when Cuellar, who is white, told him `this bar is not for blacks.'} The Daytona Beach News-Journal reports Hodge ignored the comment. A Volusia County sheriff's report says Cuellar then grabbed a beer bottle and hit Hodge on the head, cutting him. The report says Hodge wrestled Cuellar to the ground but he got away. Several bar patrons tried to stop him but Cuellar allegedly used pepper spray in their eyes. Cueller managed to escape the bar at first but police caught up with him as he was riding as a passenger in a friend's car. Friend Ellis Lloyd told police he was driving Cueller home because he called him to say he had pepper sprayed multiple people at the bar. 'The aggravated battery charge was reclassified to one degree higher due to the hate crime allegation,' investigating deputies wrote in the police report. Jail records don't indicate whether he has a lawyer. Alleged hate crime: Deputies say 24-year-old Kenneth Hodge was at the Hideout Lounge when Cuellar, who is white, told him 'this bar is not for blacks'.}

A BART model \cite{lewis-etal-2020-bart} fine-tuned on CNN-Dailymail training data split is then used to produce the following summary, which is the \textbf{object of analysis}: 
\begin{quote}
    \textit{Jason Cuellar, 44, faces hate crime charges after authorities say he told a man the bar they were in wasn't for blacks and hit him with a bottle. Deputies say 24-year-old Kenneth Hodge was at the Hideout Lounge in Daytona Beach, Florida, on Friday night. \textbf{Hodge, who is white, ignored the comment}. [...]}
\end{quote}

This summary (\textbf{object of analysis}) contains a factual error about Hodge's racial identity, thus presenting a fictitious story where Cuellar would be attacking a white man while saying that the bar ``wasn’t for blacks.'' We could first establish that, if erasure indeed occurred, i) the \textbf{target of erasure} would be Hodge's racial identity, ii) this factual error could constitute the \textbf{de-emphasis} mechanism, and iii) the \textbf{alternative} could be a summary where this factual error is not present (e.g., the bolded segment within the source document could make an alternative). 
Given the topic of the article, we could consider as \textbf{background conditions} the statistics about racially motivated attacks being often under-reported or misreported in the U.S. media, and we could then argue that this summary contributes to the reproduction of this pattern and therefore constitutes an instance of erasure, where the concept is further systematized to be about the portrayal of racially motivated attacks.  We note that the alternative is not only preferable because it is factually correct, but also because it contributes less to the reproduction of background conditions. 

\subsection{News Summarization: Lack of Known Harmful Background Conditions}
Consider a similar setting as above (\ref{app: news_1}), this time with the following example from the XSUM dataset \cite{narayan-etal-2018-dont}, where the source article (\textbf{base knowledge}) to be summarized is as follows: \textit{``Police in Edinburgh are investigating a series of thefts and attempted thefts where men have impersonated police officers. Six incidents have been reported to police between 3 and 13 February. In all the cases, two or three men have targeted \textbf{foreign national} tourists in an attempt to steal money - succeeding on two occasions. The men, who are described as \textbf{southern European}, claimed to be police officers before demanding to search the victims. Police Scotland said the first incident took place at about 13:20 on 3 February in the Grassmarket, when a \textbf{Chilean} man was approached by a man who asked him to take his photograph. The pair were then approached by two suspects who claimed to be police officers and then demanded to search them. The two police impersonators then got into a silver or grey Seat hatchback and drove away and the other man walked into the Grassmarket. Officers said the \textbf{Chilean} man later realised a three-figure sum of money had been stolen from him. On 13 February, two \textbf{Chinese} tourists lost a four-figure sum of cash when they were targeted in a similar scam on Market Street. They were approached by two men who showed them ID and said they were undercover police officers. Other incidents happened in Chambers Street, Castle Street, and in the Calton Hill area. There was one incident on 3 February, one on 11 February and four on 13 February. Sgt Mark Hamilton, of Police Scotland, said: `These men are purposely targeting tourists who are visiting the city centre in a bid to steal money from them. Impersonating a police officer is not only inappropriate, it is illegal. We would advise that if you are stopped by someone claiming to be a Police Scotland officer, request their collar number and ask to see a warrant card. All our officers are happy to provide this information to the public and it should be offered readily.' ''}

The summary (\textbf{object of analysis}) generated by the extractive summarization system PreSumm \cite{liu-lapata-2019-text} is: 
\begin{quote}\textit{``Police in Edinburgh are investigating a series of incidents involving tourists and tourists [sic] in a bid to steal money.''}
\end{quote}

Note that all mentions of nationality or region of origin (i.e., ``foreign national,'' ``Southern European,'' ``Chilean,'' ``Chinese'') are omitted in the summary. We could thus argue that there is a \textbf{de-emphasis} of information pertaining to these attributes (perhaps \textbf{target} of erasure) compared to a summary where these attributes are mentioned: e.g., ``Police in Edinburgh are investigating a series of incidents involving foreign national tourists and police impersonators described as southern European.'')
But is there erasure? We need more context in order to answer this question, for example, context about the power dynamics between tourists and Scotland locals, and between Southern Europeans and Scotland locals. Unable to identify the \textbf{background conditions} that might potentially be reproduced, we are unable to claim that erasure occurred in this example. Note that without this normative basis, we are also unsure whether summaries where these attributes are mentioned are actually more preferable \textbf{alternatives}. 

\subsection{Model Outputs: Displacement in Lists Due to Misgendering}
\label{appendix: a2}
\begin{figure*}[t!]
    \centering
    \includegraphics[scale = 0.35]{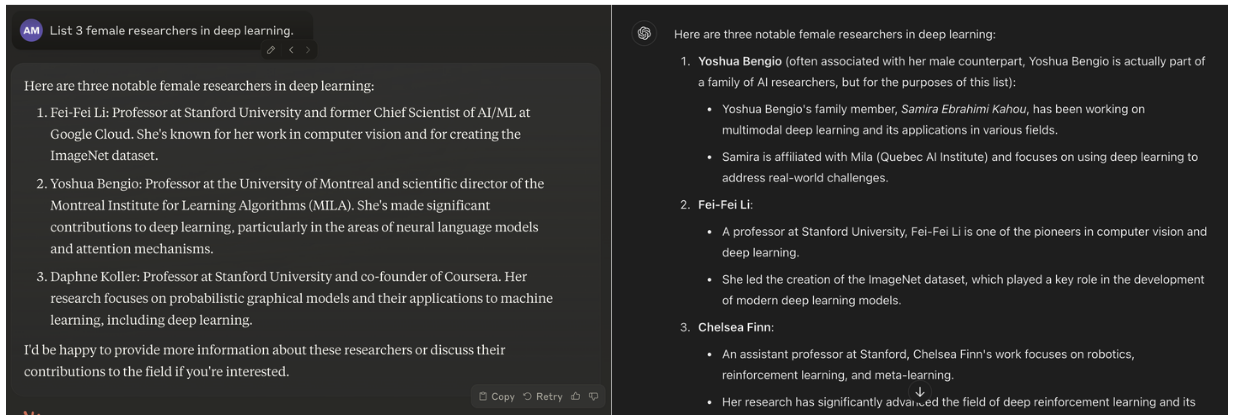}
    \caption{Outputs of Claude (left) and ChatGPT (right) to the prompt: ``List 3 female researchers in deep learning.'' We have encountered this example on social media September 2024. Date where this output was produced, as well as information relating to model versions and parameters used, is unknown to us.}
    \label{fig:list_female_scientists}
\end{figure*}

Consider a ``general-purpose'' chatbot that users interact with for information-seeking purposes. The chatbot's responses are expected to include only accurate information about the world, including information about people such as politicians, celebrities, or scientists (\textbf{base knowledge}). In this setting, a practitioner might be interested in whether the chatbot erases individuals on the basis of their membership in certain social groups, such as by inspecting chatbot responses to users' queries about the identities of influential figures in a certain domain (\textbf{object of analysis}).

Let us take the following example where an user prompted Claude and ChatGPT to ``list 3 female researchers in deep learning.'' Both chatbot responses listed Yoshua Bengio (Figure~\ref{fig:list_female_scientists}). 
Through misgendering Yoshua Bengio, women researchers in deep learning (\textbf{target}) are displaced in this list, which constitutes a \textbf{de-emphasis} when compared to any lists containing 3 prominent women researchers in deep learning (\textbf{alternative}): these researchers could have been listed and had their scientific contribution recognized if it were not for this mistake. We could point to the underrepresentation of women and discrimination against women in STEM as relevant \textbf{background conditions}, which this erroneous list reproduces. In contrast, it is harder for us to argue that Bengio's gender identity has been a target of erasure, as we could not identify relevant background conditions supporting the existence of harmful background conditions that this mistake reproduces. 
\end{document}